\documentclass{article}
\usepackage{spconf,amsmath,graphicx}
\usepackage{multicol,multirow}
\usepackage{microtype}
\usepackage{graphicx}
\usepackage{subfigure}
\usepackage{booktabs} 
\usepackage{float}
\usepackage{color}

\title{Visual Explanation via Similar Feature Activation for Metric Learning}
\name{
{Yi Liao}$^{\star}$, {Ugochukwu Ejike Akpudo}$^{\star}$, {Jue Zhang}$^{\star}$, 
{Yongsheng Gao}$^{\star}$, {Jun Zhou}$^{\star}$,
{Wenyi Zeng}$^{\dagger}$, {Weichuan Zhang}$^{\dagger}$
}
\address{$^{\star}$Griffith University\\
$^{\dagger}$Shaanxi University of Science and Technology
}

\begin{document}
%
\maketitle
\begin{abstract}
Visual explanation maps enhance the trustworthiness of decisions made by deep learning models and offer valuable guidance for developing new algorithms in image recognition tasks. Class activation maps (CAM) and their variants (e.g., Grad-CAM and Relevance-CAM) have been extensively employed to explore the interpretability of softmax-based convolutional neural networks, which require a fully connected layer as the classifier for decision-making. However, these methods cannot be directly applied to metric learning models, as such models lack a fully connected layer functioning as a classifier. To address this limitation, we propose a novel visual explanation method termed Similar Feature Activation Map (SFAM). This method introduces the channel-wise contribution importance score (CIS) to measure feature importance, derived from the similarity measurement between two image embeddings. The explanation map is constructed by linearly combining the proposed importance weights with the feature map from a CNN model. Quantitative and qualitative experiments show that SFAM provides highly promising interpretable visual explanations for CNN models using Euclidean distance or cosine similarity as the similarity metric.
\end{abstract}
\section{Introduction}
\label{sec:intro}
Deep neural networks, particularly convolutional neural networks (CNN), have led to unprecedented breakthroughs in image classification tasks. Most image classification models require fully connected (FC) layers as classifiers that are trained via supervised learning, which requires annotated training datasets for a specific task. They make the classification decision by the highest probability in all classes. The probability is calculated via the softmax function, which is heavily dependent on the FC layer (classifier). These models,  commonly known as softmax-based CNNs, can effectively recognize a testing sample from the class domain of the training set but struggle to identify a testing image from the unseen class. Unfortunately, it is impossible to collect the training images that cover endless classes, necessitating the studies on metric learning classification models that attract attention in computer vision community. Metric learning models can accurately recognize novel classes that are disjoint from the training set by making classification decisions based on similarity measurements rather than a linear classifier. Metric learning has proven effective for tasks such as few-shot learning~\cite{FRN} and image retrieval~\cite{MultiSimilarity}.
\begin{figure}[t]
\begin{minipage}[b]{1.0\linewidth}
\centering
\centerline{\includegraphics[width=8.5cm]{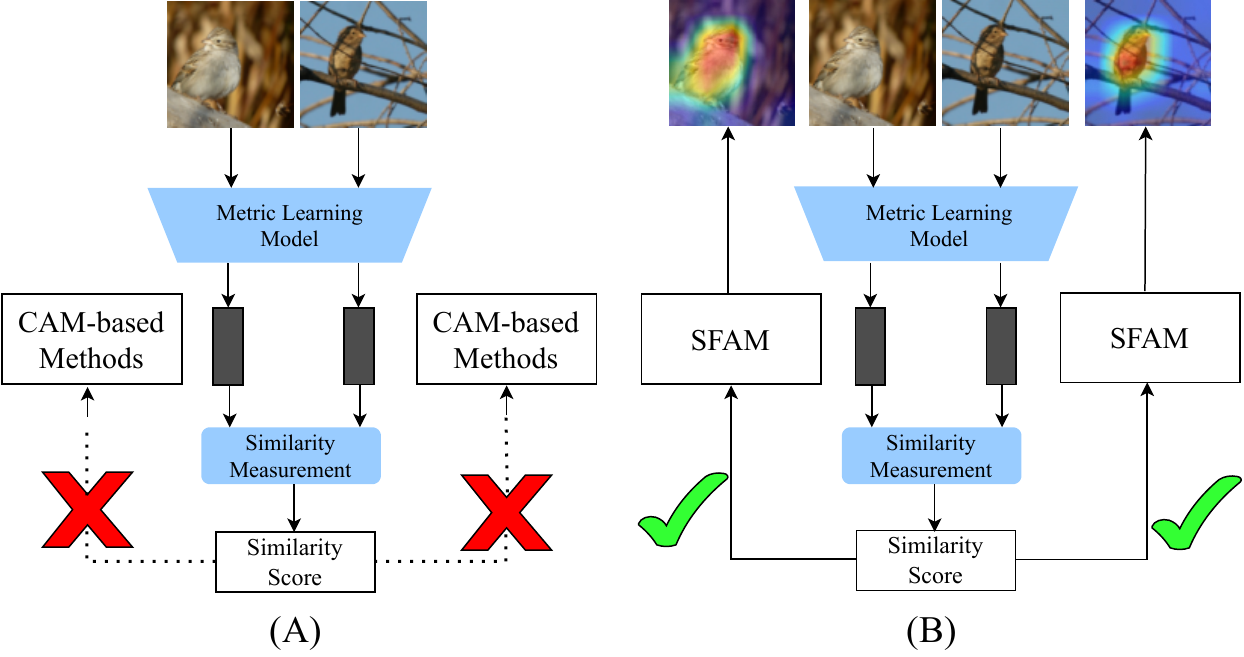}}
\end{minipage}
\caption{(A). CAM-based methods(e.g.,CAM~\cite{CAM}, Grad-CAM~\cite{GradCAM}, and Relevance-CAM~\cite{RelevanceCAM}) can not be used  for interpreting metric-learning CNNs because of no FC layer as the classifier. (B). The proposed SFAM can generate the explanation map for the metric learning model, which highlights the body of the two birds as the similar feature for decision-making.}
\label{fig:problem}
\vspace{-0.6 cm}
\end{figure}

Visual explanation methods aim at interpreting image classification models by highlighting where the model looked, i.e. the contributing pixels on an input image that predict the class label. However, much less work has focused on exploring the visual explanation for metric learning models. Current widely used explanation methods, including, but not limited to, RISE~\cite{RISE}, Class Activation Map (CAM)~\cite{CAM},  GradCAM-based methods~\cite{GradCAM,GradCAM++,XGradCAM,LayerCAM}, Relevance-CAM~\cite{RelevanceCAM}, LIFT-CAM~\cite{LIFTCAM}, and Ablation-CAM~\cite{AblationCAM}, are developed to interpret softmax-based CNNs. As shown in Fig.~\ref{fig:problem} (A), the absence of a FC layer in these CNNs' classifiers limits these visual explanation methods for metric learning~\cite{GradFAMScoreFAM}. Against the conventional use of Euclidean distance as the similarity metric, state-of-the-art visual explanation methods for metric learning ~\cite{Decomposition,RAM,CG-RAM} predominantly employ the cosine similarity metric for CNN explanation. However, visualizing the decision-making regions on input images for a metric-agnostic CNN models remains a challenge. 

To solve this problem, we propose a novel visual explanation algorithm called Similar Feature Activation Map (SFAM). The proposed SFAM aims to highlight discriminative image pixels/regions that contribute to a metric learning model's decision-making logic. To this end, we propose a channel-wise contribution importance score (CIS) that assigns a weight to each channel. A class prediction can be explained by assigning a weight to each feature, indicating its influence, with the key to the explanation being the contribution of individual features. Specifically, the comparison between two vectors representing two input images respectively, the closer the two elements from the same channel are, the more important the channel will be. A single convolution kernel is responsible for capturing an individual feature, and the contribution of an arbitrary channel can represent the contribution of that feature. The proposed CIS, serving as the importance coefficients, can be linearly combined with the feature map out of the final layer of a CNN model to create the SFAM. Hence, the proposed SFAM highlights similar features by awarding a higher value, as shown in Fig.~\ref{fig:problem} (B). In a word, the key contribution in this work can be summarized as follows,
\begin{itemize}
\item We propose a novel interpretable algorithm SFAM for generating visual explanation maps that highlight similar features looked at by metric-learning CNN models in few-shot learning image classification and image retrieval tasks.
\item The proposed SFAM has a flexible architecture, allowing the use of cosine similarity metric or Euclidean distance metric, unlike existing methods, which typically rely on cosine similarity for metric learning.  
\item The proposed SFAM provides visual explanations for any metric-learning CNN without any constraints to the CNN architecture.
\item The quantitative and qualitative experiments demonstrate the effectiveness of the proposed method in few-shot image classification and image retrieval tasks.
\end{itemize}
\section{Related Work}
\label{sec:relatedwork}
\subsection{Visual Explanation for Softmax-based CNN Models}
\label{sec:relatedworksub1}
The previous works~\cite{SailencyMap,GuidedBP} attempt to interpret CNNs by generating the saliency maps, which highlights the contributing pixels relevant to the prediction of a specific class. As the first effort to interpret CNNs by highlighting the contributing regions on an input image, CAM~\cite{CAM} is constructed by linearly combining the output feature map and the weights from the FC layer (classifier). Grad-CAM~\cite{GradCAM}, Grad-CAM++~\cite{GradCAM++}, XGradCAM~\cite{XGradCAM}, and LayerCAM~\cite{LayerCAM} all use the gradient of the classification score $w.r.t.$ the neurons to generate the explanation map. Ablation-CAM~\cite{AblationCAM} compute the importance of the feature map by using the variation of classification score obtained via multiple iteration. Relevance-CAM~\cite{RelevanceCAM} inevitably requires the number of classes to calculate the initial relevance score when interpreting a CNN. RISE~\cite{RISE} interprets CNNs by randomized perturbations on input image. However, the above explanation methods are all built on the classification score, the computation of which relies heavily on the FC layer as the classifier. Therefore, they cannot be adopted to interpret metric-learning CNN models. 

\subsection{Visual Explanation for Metric Learning}
\label{sec:relatedworksub2}

The aforementioned explanation methods for softmax-based CNNs are ineffective for metric-learning CNNs, which do not utilize a fully connected (FC) layer as a classifier for decision-making. To address this limitation, several methods have been developed, including Decomposition~\cite{Decomposition}, RAM~\cite{RAM}, CG-RAM~\cite{CG-RAM}, and Grad-FAM~\cite{GradFAMScoreFAM}, to generate visual explanations. Decomposition~\cite{Decomposition} interprets the CNN models that use the cosine metric as a similarity measure, which restricts its general applicability across different similarity metrics. Ranking Activation Map (RAM)~\cite{RAM} generates an explanation map by calculating the importance of each channel based on the feature representation of the input image, highlighting features within the image without focusing on similarities between different images. RAM, unfortunately, highlights the features that do not necessarily correspond to similarities between two images, necessitating Confidence Gradient-weighted RAM (CG-RAM)~\cite{CG-RAM} which improves RAM by using the gradient of the cosine similarity value to determine channel importance. Consequently, CG-RAM often highlights background regions irrelevant to the object features, reducing its effectiveness. Grad-FAM~\cite{GradFAMScoreFAM} focuses solely on the given image and does not effectively highlight similar features between two images, limiting its utility in comparative tasks. 

\begin{figure*}[t]
\centering
\includegraphics[width=1.0\textwidth]{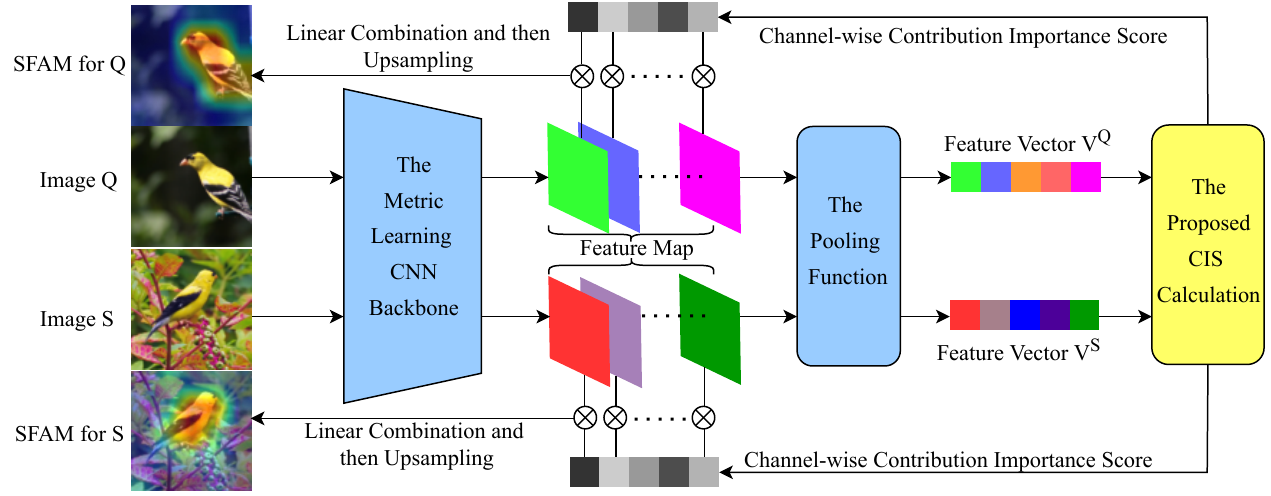}
\caption{The pipeline of the proposed SFAM. Image Q and Image S are fed into a metric learning CNN backbone for extraction of the feature maps. The feature maps are embedded into the feature vectors, which are utilized to calculate the proposed CIS. The proposed SFAMs for Image Q and Image S are generated by linearly combining the feature maps and the corresponding CIS.}
\label{fig:pipeline}
\vspace{-0.5 cm}
\end{figure*}
\section{The Proposed Method}
\label{sec:method}
The pipeline for the proposed Similar Feature Activation Map (SFAM) is illustrated in Fig.~\ref{fig:pipeline}. As shown in Fig.~\ref{fig:pipeline}, two images are fed into a metric-learning CNN backbone for the extraction of feature maps. Then, the two feature maps are embedded into the two feature vectors. The explanation maps generated by the proposed SFAM are constructed by linearly combining the feature maps and the proposed channel-wise CIS. The proposed CIS algorithm can be applied to Euclidean distance or cosine similarity metrics. 


\subsection{Channel-wise Contribution Importance Score}
\label{subsec:cis}
Let $f(\cdot)$ to denote a CNN backbone network for metric learning.  
$Q$ and $S^c$ denote a query image and a support image from the specific class $c$, respectively. Thus, $f(Q)$ and $f(S^c)$ represents the feature map from the query image and the support image respectively, so $f(Q)$, $f(S^c)$ $\in  R^{N\times H \times W}$ where $N$, $H$, and $W$ denote the number of channels, height and width of the feature map respectively. Let $p(\cdot)$ to denote a pooling function. The query  representation $V^Q$ (feature vector) can be denoted by $p(f(Q)) \in R^{N}$. Suppose that there are $K$ support images from the class $c$, the support representation $V^S$ can be expressed by 
\begin{equation}
 \centering
 \label{prototype}
 V^S=\frac{1}{K}{\sum_{k=1}^Kp(f(S^c_k))}.
\end{equation}
$V^Q$ and $V^S$ represent the feature vectors of $N$ elements from the query image $Q$ and the support image $S$ respectively. Let $V^Q=[v^Q_1,v^Q_2,\cdots,v^Q_N]$ and $V^S=[v^S_1,v^S_2,\cdots,v^S_N]$, the scalar $v^Q_n$  and $v^S_n$ are the elements at the $n$-th channel of $V^Q$ and $V^S$ respectively and $n=1,\cdots,N$. The $n$-th channel CIS ${\omega_{n}}$ can be calculated by the following equation,
\begin{equation}
\centering
\label{CIS}
\omega_n=\frac{1}{N-1}-\frac{(v^Q_n-v^S_n)^2}{(N-1)\cdot \sum_{m=1}^N(v_m^Q-v_m^S)^2},
\end{equation}
where $N$ denotes the number of channels of the feature map out of the final layer from a CNN backbone.
\subsection{Similar Feature Activation Map}
\label{subsec:sfam}
Let $A_{i,j}^n$ to denote the value of neuron at location $(i,j)$ on the $n$-th channel of the feature map from the final layer of a CNN backbone. The proposed similar feature activation map (SFAM) is defined by
\begin{equation}
 \centering
 \label{SFAM}
L_{\text{SFAM}}=\sum_{n=1}^N{\text{Norm}(\omega_n) \cdot A_{i,j}^n},
\end{equation}
where $\text{Norm}(\cdot)$ is max-min normalization function mapping score within $[0,1]$ by 
\begin{equation}
\centering
\label{max-min-norm}
\text{Norm}(\omega_n)=\frac{\omega_n-\text{Min}(\{ \omega_n \}_{n=1}^N )}{\text{Max}(\{ \omega_n \}_{n=1}^N)-\text{Min}(\{ \omega_n \}_{n=1}^N)}.
\end{equation}
From Eq.~\ref{SFAM}, the visualization of the proposed $L_{\text{SFAM}}$ is generated by using bilinear interpolation as an up-sampling technique.
\subsection{Visual Explanation for CNN using Cosine Similarity}
\label{subsec:cis_cosine}
Let $\hat{V}^Q=[\hat{v}_1^Q, \hat{v}_2^Q, \cdots,\hat{v}_N^Q]$ and $\hat{V}^S=[\hat{v}_1^S,\hat{v}_2^S,\cdots, \hat{v}_N^S]$ to denote the $L2$ normalization of $V^Q$ and $V^S$ respectively. Hence, $\hat{V}^Q=\frac{V^Q}{||V^Q||}$ and $\hat{V}^S=\frac{V^S}{||V^S||}$ where $\parallel \cdot \parallel$ indicates the length of a vector. By respectively replacing $v_n^Q$ and $v_n^S$ with $\hat{v}_n^Q$ and $\hat{v}_n^S$ in Eq.~\ref{CIS}, we can obtain the following score $\hat{\omega}_n$, 
\begin{equation}
\label{cosine_adaption}
\begin{aligned}
\hat{\omega}_n=\frac{1}{N-1}-\frac{(\hat{v}^Q_n-\hat{v}^S_n)^2}{(N-1)\cdot{\sum_{m=1}^N{(\hat{v}^Q_m-\hat{v}^S_m)}^2}}~.
\end{aligned}
\end{equation}
Next, the proposed SFAM for cosine similarity metric can be obtained by replacing $\omega_n$ in Eq.~\ref{SFAM} with $\hat{\omega}_n$ defined in Eq.~\ref{cosine_adaption},
\begin{equation}
\centering
 \label{SFAM_cosine}
L_{\text{SFAM}}^{\text{cos}}=\sum_{n=1}^N{\text{Norm}(\hat{\omega}_n) \cdot A_{i,j}^n}. 
\end{equation}
When the similarity between two image representations $V^Q$ and $V^S$ is measured via cosine metric, the value of the proposed $\hat{\omega}_n$ is positively correlated with $\hat{v}_n^Q\cdot\hat{v}_n^S$. Hence, $\hat{\omega}_n$ can be used to measure the contribution to cosine value.\\

\noindent \textit{Proof}: Let $\text{cos}(\cdot)$ to denote the cosine function, from Eq.~\ref{cosine_adaption}, we can infer the following equation, 
\begin{equation}
\label{cosine}
\begin{aligned}
\hat{\omega}_n
=&\frac{{\sum_{m \ne n} {(\hat{v}^Q_m)}^2}+{\sum_{m \ne n}{(\hat{v}^S_m)}^2}-2\text{cos}(V^Q,V^S)}{(N-1)(2-2\text{cos}(V^Q,V^S))}\\
+&\frac{\hat{v}^Q_n \cdot \hat{v}^S_n}{(N-1)(1-\text{cos}(V^Q,V^S))}.
\end{aligned}
\end{equation}
 According to Eq.~\ref{cosine}, when the decision is made, the value of $\text{cos}(V^Q,V^S)$ is fixed. Assume $\text{cos}(V^Q,V^S) \neq 1$, then $1-\text{cos}(V^Q,V^S) >0$ and $(N-1)>0$ because $V^Q$ and $V^S$ are vectors. The value of $\hat{\omega}_n$ increases as the value of $\hat{v}_n^Q\cdot\hat{v}_n^S$ increases. $\hat{v}_n^Q\cdot\hat{v}_n^S$ represents the $n$-th channel contribution to $\text{cos}(V^Q,V^S)$ because $\text{cos}(V^Q,V^S)=\sum_{n=1}^N{\hat{v}_n^S \cdot \hat{v}_n^Q}$. Therefore, $\hat{\omega}_n$ can be used to measure the $n$-th channel contribution to $\text{cos}(V^Q,V^S)$.
\begin{figure}[t]
\begin{minipage}[b]{1.0\linewidth}
\centering
\centerline{\includegraphics[width=8.5cm]{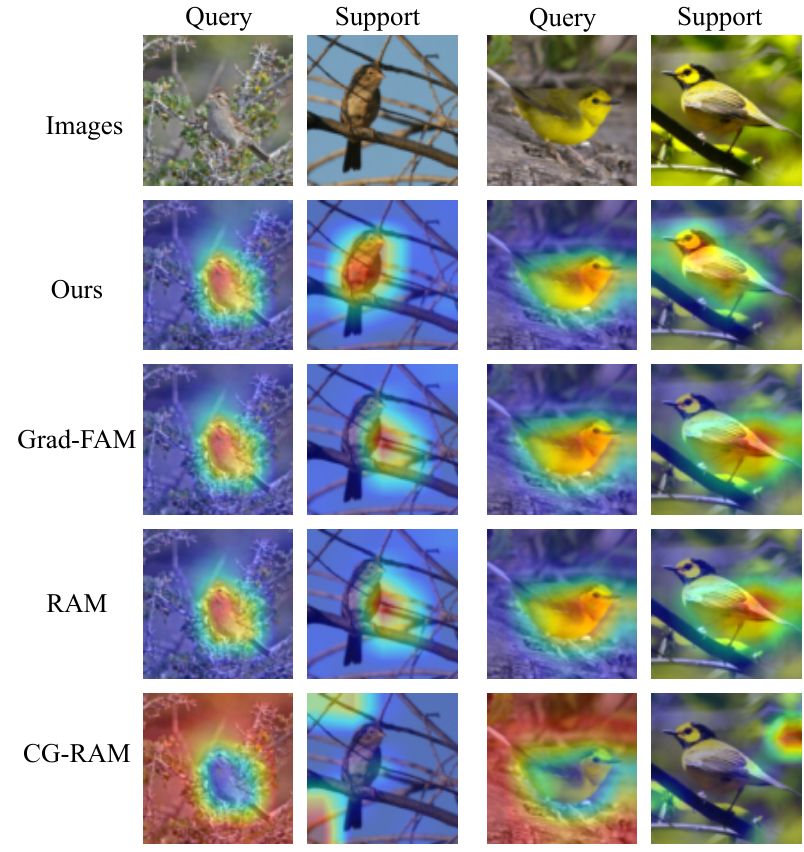}}
\end{minipage}
\caption{The qualitative comparison between the proposed SFAM and $3$ baseline methods for FRN (ResNet12) that uses Euclidean distance as the similarity metric in $5$-way $1$-shot image classification task.}
\label{fig_EuclideanDistance}
\vspace{-0.5 cm}
\end{figure}

\section{Experiment}
\label{sec:Experiment}
\subsection{Dataset and Implementation Details}
\label{ssec:dataset}
The qualitative and quantitative experiments are conducted on widely-used fine-grained image dataset CUB200~\cite{CUB200}, which includes $11,788$ images over $200$ classes. For few-shot image classification, the split strategy is the same as in~\cite{SplitCUB}. The $200$ classes are splited into $100$, $50$, and $50$ for training, validation and testing, respectively. For image retrieval tasks, by following the same split strategy as in~\cite{MultiSimilarity}, the first $100$ classes are used for training and the remaining $100$ classes with $5,924$ images are used for testing set. The evaluation of the proposed SFAM for both tasks are performed on the testing set. It is worth noting that the testing class domain is disjoint with the training set. The testing set provides the bounding box annotations which are used for evaluating localization capacity. Feature Map Reconstruction Network (FRN)~\cite{FRN} and General Pair Weighting framework (GPW)~\cite{MultiSimilarity} are selected to evaluate the effectiveness of the proposed method. FRN makes the decision via Euclidean distance as the similarity metric for few-shot image classification and GPW uses cosine similarity as the similarity metric for image retrieval. We implement the publicly released code\footnote{https://github.com/Tsingularity/FRN} to obtain FRN model and the released code\footnote{https://github.com/MalongTech/research-ms-loss} to obtain GPW model. ResNet12 and ResNet50 are employed as the CNN backbone to extract the feature map for FRN and GPW respectively. All experiments are implemented by using Pytorch library with Python 3.8 on NVDIA RTX3090 GPU.
\begin{figure}[t]
\begin{minipage}[b]{1.0\linewidth}
\centering
\centerline{\includegraphics[width=8.5cm]{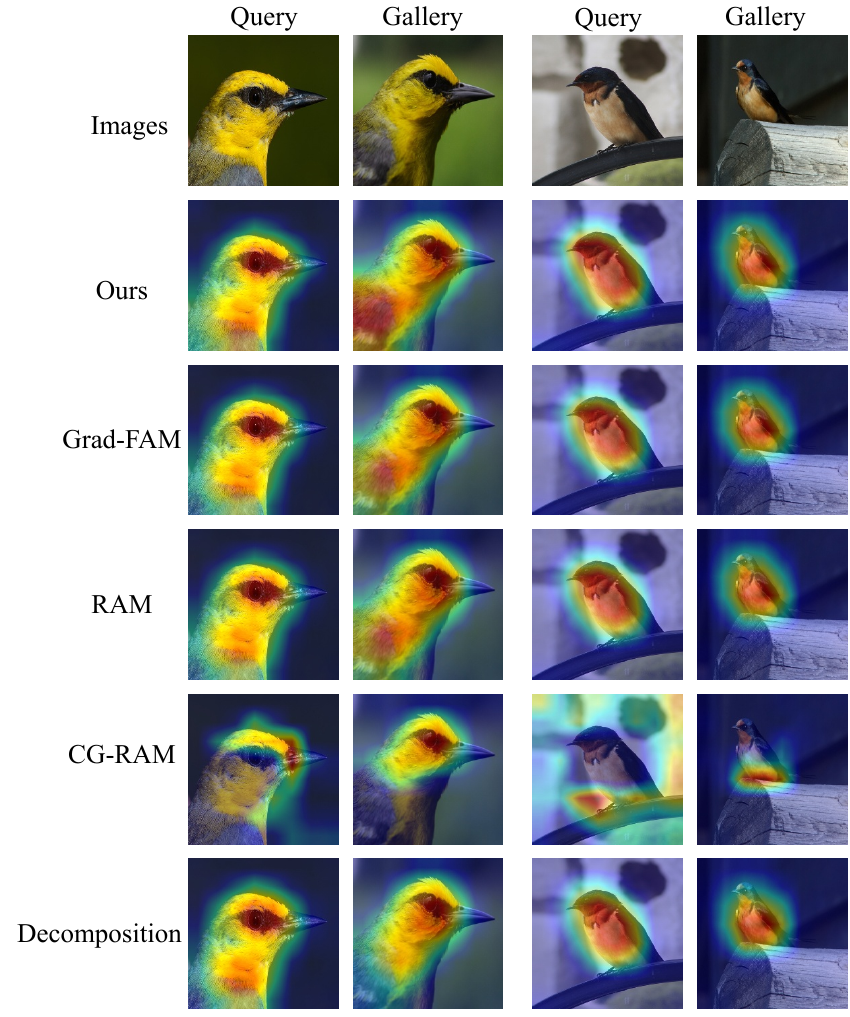}}
\end{minipage}
\caption{The qualitative comparison between the proposed SFAM and $4$ baseline methods for GPW (ResNet50) that uses cosine as the similarity metric in image retrieval task.}
\label{fig_CosineSimilarity}
\vspace{-0.6 cm}
\end{figure}
\subsection{Qualitative Experiment}
\label{ssec:qualitative}
To evaluate the effectiveness of the proposed SFAM, the qualitative comparison between SFAM and $4$ explanation methods (e.g.,Decomposition, RAM, CG-RAM, and Grad-FAM) are conducted on the same images for FRN (ResNet12) and GPW (ResNet50). It is worth noting that Decomposition fails to interpret CNN model that using Euclidean Distance as the metric. Fig.~\ref{fig_EuclideanDistance} illustrates the comparison of $4$ images for FRN (ResNet12) in $5$-way $1$-shot classification. As shown in Fig.~\ref{fig_EuclideanDistance}, the proposed SFAM can correctly highlight the similar features (the birds' body between the left query and left support images and the bird's head between the right query and support images). However, the regions highlighted by RAM, GradFAM, and CG-RAM are not the similar features between two birds actually. Fig.~\ref{fig_CosineSimilarity} visualizes the comparison of $4$ samples for GPW (ResNet50) in image retrieval. As shown in Fig.~\ref{fig_CosineSimilarity}, the proposed SFAM can successfully generate the explanation maps for metric-learning CNN model using cosine as the similarity metric.
\begin{table}[t]
\caption{The quantitative comparison of localization capacity between the proposed SFAM and $4$ baselines for FRN (ResNet12) using Euclidean Distance as the similarity metric. Higher value is better for average IoU and Accuracy (IoU$\ge 0.5$)}
 \label{table_FRN}
\begin{center}
\begin{tabular}{ccc}
\toprule
       \multirow{2}{*}{FRN (ResNet12)} & \multicolumn{2}{c}{Euclidean Distance}\\
       \cline{2-3}
        & IoU (\%)& Accuracy(\%) \\
\midrule
    CG-RAM~\cite{CG-RAM}&$24.13$& $16.43$\\
    RAM~\cite{RAM}&   $49.85$& $61.24$\\
    Grad-FAM~\cite{GradFAMScoreFAM}&$49.72$&$62.31$\\
    Decomposition~\cite{Decomposition}&-&-\\
    Ours&$\boldsymbol{52.33}$&$\boldsymbol{69.74}$ \\
\bottomrule
\end{tabular}
\end{center}
\vspace{-0.5 cm}
\end{table}


\subsection{Quantitative Experiment}
\label{ssec:quantitative}
To verity the localization capacity of the proposed SFAM, the quantitative experiments are conducted on FRN (ResNet12) and GPW(ResNet50) to compare our method with $4$ baseline methods (e.g., Decomposition, RAM, CGRAM, and GradFAM). IoU(intersection over Union)~\cite{GradCAM,LayerCAM,Decomposition} and localization accuracy (IoU $\geq 0.5$)~\cite{LayerCAM,Decomposition} are adopted to measure the localization capacity. The threshold is set to $0.2$ in our experiment.
Average IoU is calculated over all the correctly predicted samples. Specifically, in few-shot image classification, $5$-way $1$-shot is chosen for comparison, the average IoU is calculated over $10,000$ episodes. Table~\ref{table_FRN} and Table~\ref{table_GPW} show the quantitative comparison result. From the results, the proposed SFAM achieves the best performance in average IoU and localization accuracy, outperforming other methods by at least $2.2\%$.
\begin{figure}[t]
\begin{minipage}[b]{1.0\linewidth}
\centering
\centerline{\includegraphics[width=8.5cm]{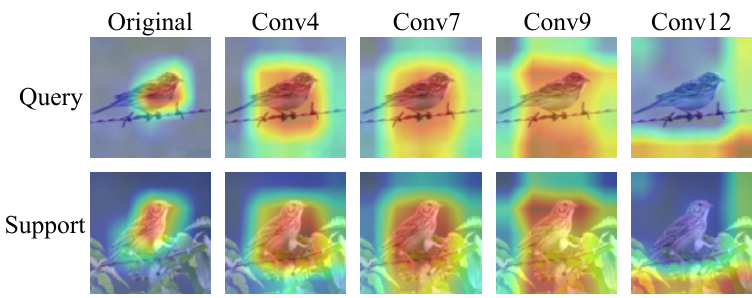}}
\end{minipage}
\caption{Sanity check for the proposed SFAM by cascading randomization from the $1$-st convolution layer to the $4$-th, the $7$-th, the $9$-th, and the $12$-th convolution layer in FRN (ResNet12) respectively. The original explanation map is generated by the propsoed SFAM for the well-trained FRN (ResNet12) without any parameters randomization.}
\label{fig_sanitycheck}
\vspace{-0.4 cm}
\end{figure}
\begin{table}[t]
\caption{The quantitative comparison of localization capacity between the proposed SFAM and $4$ baselines for GPW (ResNet50) using Cosine as the similarity metric. Higher value is better for average IoU and Accuracy (IoU$\ge 0.5$).}
 \label{table_GPW}
\begin{center}
\begin{tabular}{ccc}
\toprule
       \multirow{2}{*}{GPW (ResNet50)} & \multicolumn{2}{c}{Cosine Similarity}\\
       \cline{2-3}
        & IoU (\%)& Accuracy(\%) \\
\midrule
    CG-RAM~\cite{CG-RAM}&$25.16$& $20.23$\\
    RAM~\cite{RAM}&   $52.97$& $60.13$\\
    Grad-FAM~\cite{GradFAMScoreFAM}&$53.38$&$60.34$\\
    Decomposition~\cite{Decomposition}&$52.91$&$58.77$\\
    Ours&$\boldsymbol{55.65}$&$\boldsymbol{69.08}$ \\
\bottomrule
\end{tabular}
\end{center}
\vspace{-0.5 cm}
\end{table}

\subsection{Sanity Check}
\label{ssec:sanity check}
Sanity check aims to evaluate the sensitivity of the explanation maps to the parameters of the deep learning models~\cite{SanityCheck}. We evaluate the proposed SFAM by cascading randomizing the paramters' weights of FRN (ResNet12). Fig.~\ref{fig_sanitycheck} shows that the sanity analysis for the explanation map from output layer of ResNet12. The results show that the explanation maps are unfaithful with the parameters randomization, revealing the poor localization performance by SFAM on the image. This verifies that the proposed SFAM is sensitive to the changes in CNN parameters. This indicates that SFAM depends on the well-trained weights of the CNN for generating accurate explanations. Furthermore, the results emphasize the need for well-trained models to ensure reliable visual explanations.
\section{Conclusion}
\label{sec:conclusion}
This work proposes SFAM, a novel visual explanation method designed for metric-learning CNNs. SFAM addresses the challenge of interpreting metric-learning models, which lack the fully connected layer found in traditional softmax-based CNNs. Unlike existing methods that primarily rely on cosine similarity, SFAM offers flexibility by supporting both Euclidean distance and cosine similarity metrics, offering a channel-wise CIS metric for feature importance estimations. Qualitative and quantitative experiments on the CUB200 dataset demonstrate the effectiveness of SFAM in few-shot image classification task, outperforming existing visual explanation methods.




\vfill\pagebreak

\bibliographystyle{IEEEbib}
\bibliography{SimilarFeatureActivationMap}

\end{document}